\documentclass[letterpaper, 10 pt, conference]{ieeeconf}  

\IEEEoverridecommandlockouts                              

\usepackage[acronym]{glossaries}
\usepackage{graphicx}
\usepackage{amsfonts}
\usepackage{siunitx}
\usepackage{booktabs}
\usepackage[export]{adjustbox}
\usepackage{xcolor}
\usepackage{algorithm}
\usepackage{subfig}
\usepackage{xspace}
\usepackage{algpseudocode}
\usepackage{makecell}
\usepackage{hyperref}

\definecolor{todo-red}{RGB}{200,12,12}
\definecolor{green4}{RGB}{0,128,0}

\newcommand{\acronym}[1]{\gls{#1}\@}
\newcommand{\acronymfirst}[1]{\glsfirst{#1}\@}

\newacronym{fpfh}{FPFH}{Fast Point Feature Histogram}
\newacronym{icp}{ICP}{Iterative Closest Point}
\newacronym{iss}{ISS}{Intrinsic Shape Signature}
\newacronym{pca}{PCA}{Principal Component Analysis}
\newacronym{csg}{CSG}{Constrained Similarity Graph}
\newacronym{crf}{CRF}{Conditional Random Field}
\newacronym{board}{BOARD}{BOrder Aware Repeatable Directions}
\newacronym{tsdf}{TSDF}{Truncated Signed Distance Field}
\newacronym{ros}{ROS}{Robot Operating System}
\newacronym{gsm}{GSM}{Global Segmentation Map}
\newacronym{slam}{SLAM}{Simultaneous Localization and Mapping}
\newacronym{rmse}{RMSE}{Root-mean-square error}
\newacronym{ransac}{RANSAC}{Random Sample Consensus}

\makeatletter

\DeclareRobustCommand\onedot{\futurelet\@let@token\@onedot}
\def\@onedot{\ifx\@let@token.\else.\null\fi\xspace}

\def\eg{\emph{e.g}\onedot}

\makeatother

\begin{document}
%

\title{Incremental Object Database: \\ Building 3D Models from Multiple Partial Observations}
%
%
%

\author{Fadri~Furrer,
        Tonci~Novkovic,
        Marius~Fehr,
        Abel~Gawel,
        Margarita~Grinvald,\\
        Torsten~Sattler,
        Roland~Siegwart,
        and~Juan~Nieto
\thanks{F. Furrer, T. Novkovic, M. Fehr, A. Gawel, M. Grinvald, R. Siegwart, and J. Nieto are with the Autonomous Systems Lab, ETH, 8092 Zurich, Switzerland, e-mail: \{fadri.furrer, tonci.novkovic, marius.fehr, abel.gawel, margarita.grinvald\}@mavt.ethz.ch, \{rsiegwart, nietoj\}@ethz.ch.}%
\thanks{T. Sattler is with the Computer Vision Group, Deparment of Computer Science, ETH, 8092 Zurich, Switzerland, e-mail: sattlert@inf.ethz.ch.}
\thanks{Find the accompanied video here: \url{https://youtu.be/9_xg92qqw70}.}}

\maketitle

\begin{abstract}
Collecting 3D object datasets involves a large amount of manual work and is time consuming.
Getting complete models of objects either requires a 3D scanner that covers all the surfaces of an object or one needs to rotate it to completely observe it.
We present a system that incrementally builds a database of objects as a mobile agent traverses a scene.
Our approach requires no prior knowledge of the shapes present in the scene.
Object-like segments are extracted from a global segmentation map, which is built online using the input of segmented RGB-D images.
These segments are stored in a database, matched among each other, and merged with other previously observed instances.
%
%
%
This allows us to create and improve object models on the fly and to use these merged models to reconstruct also unobserved parts of the scene.
%
%
%
The database contains each (potentially merged) object model only once, together with a set of poses where it was observed.
We evaluate our pipeline with one public dataset, and on a newly created Google Tango dataset containing four indoor scenes with some of the objects appearing multiple times, both within and across scenes.
\end{abstract}


%
\IEEEpeerreviewmaketitle

\section{Introduction}
\begin{figure}[!t]
\centering
\includegraphics[width=\columnwidth]{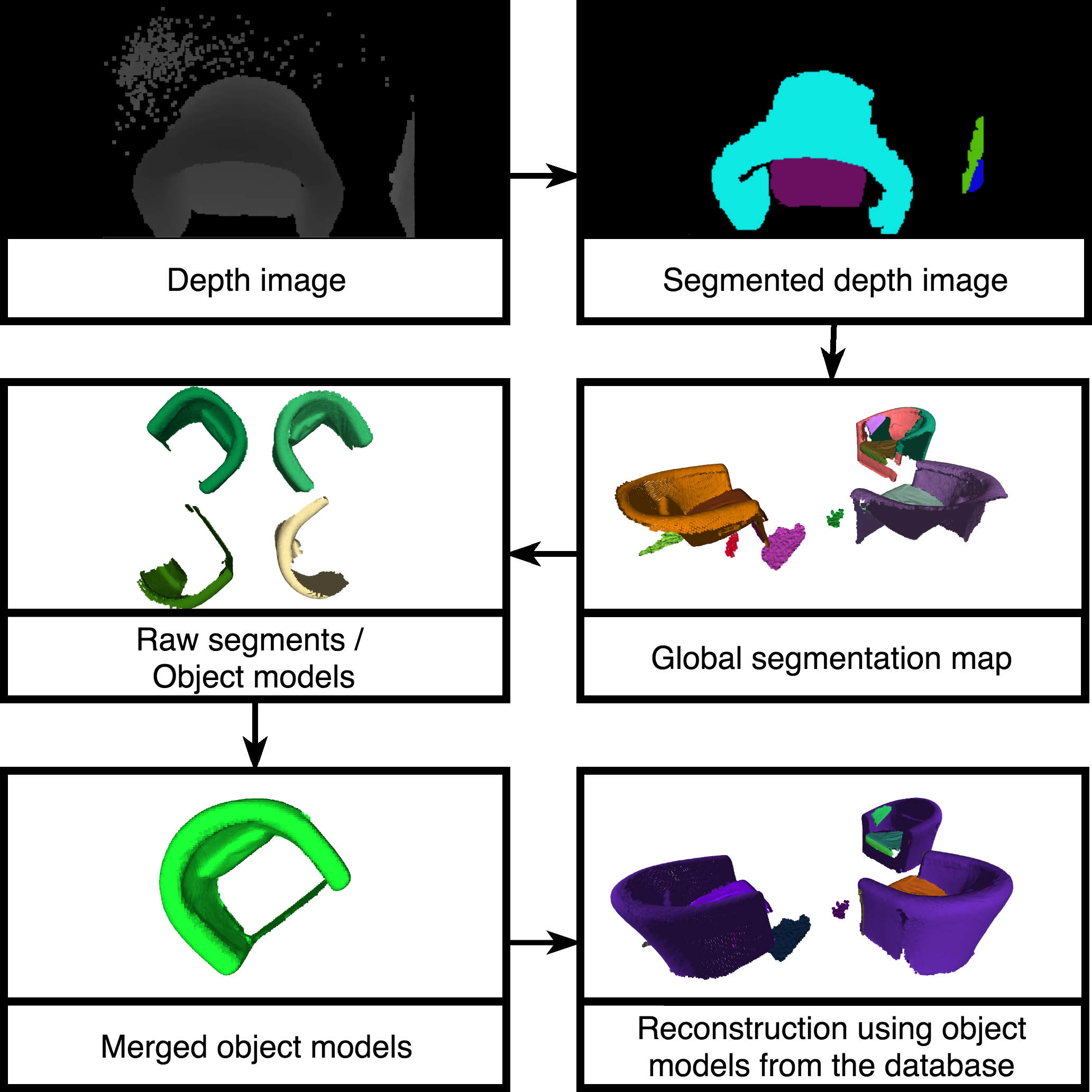}
\caption{Input depth images, top left, get segmented using a geometrical segmentation (\ref{sec:depth_segmentation}), the individual depth segments, top right, are integrated into a \acronym{gsm} (\ref{sec:gsm}), middle right, from which we can extract raw segments. We put these raw segments in a database of object models and describe their keypoints (\ref{sec:keypoints_and_descriptors}), middle left, and try to match and merge them with other models built from observations of instances of the same object kind (\ref{sec:matching_and_registration},~\ref{sec:merging}), bottom left. With these merged object models, we can reconstruct the scene and complete it by combining the data of all observed object instances, bottom right. Different colors indicate different segments.}
\label{fig:teaser_image}
\end{figure}

Humans have an excellent understanding of an environment's structure.
Once we observe the objects in a scene, we efficiently memorize them and can easily recall the complete shape of these objects regardless of occlusions or only partial observations.
We can furthermore apply the knowledge of repeating objects to immediately recognize them in new scenes, and even hallucinate their full shape despite partial observations.
However, modern robotic systems usually lack a notion of objects and merely work based on their sensor data, e.g., storing scenes as abstract point clouds.
This can be highly inefficient for scene reconstructions, as any structure needs to be observed from multiple view-points and leads to large map sizes.
During the reconstruction process, repetitive structures typically result in ambiguous data associations.
Furthermore, such systems exhibit a lack of scene understanding, creating a need for post-processing steps that perform object recognition.
Contemporary approaches for 3D object recognition require databases of pre-recorded objects to be able to match these to the scene data.
Typically, the creation of these databases involves significant human labor in terms of either hand-modeled 3D meshes \cite{li2015database} or 3D models obtained via careful manual scanning \cite{salas2013slam++}.

In this work, we present an automatic system to incrementally build a database of 3D object models using depth sensing.
Rather than aiming to detect a fixed set of objects in the RGB images provided by an RGB-D sensor \cite{russakovsky2012object, deng2014large, simonyan2013deep}, our approach segments the depth image provided by the sensor based on an edge and convexity map similar to \cite{uckermann2013realtime, tateno2015real}.
In subsequent steps, the initial depth segments are integrated into a \acronym{tsdf} stored in a voxel grid resulting in a \acronymfirst{gsm}.
From this \acronym{gsm}, raw segments are extracted if the corresponding voxels are untouched for a certain time, described using features, and inserted in an object database as object models.
In the database they are matched to existing object models, which get refined and completed over time.
By re-projecting the merged object models to all locations where the object instances were observed a scene can be completed.
This process is illustrated in Fig.~\ref{fig:teaser_image}.

One of the advantages of our approach is that we do not need to pre-define which objects we want to detect, rather, we automatically detect objects based on a segmented scene reconstruction.
This is a prerequisite for building a system that makes scanning objects as simple as walking through an everyday scene, without the need to place the objects in front of a scanning setup, e.g. a turn-table.
Our method can be used to construct 3D maps consisting of planar segments, which are geometrically not distinctive, and a set of (potentially incomplete) object models.
%
%
Since these maps are constructed incrementally, our approach is directly applicable to scenarios where a robot (or another type of autonomous agent) needs to interact with the world on an object level in order to solve its tasks.

This paper makes the following contributions:
\begin{itemize}
    \item A novel approach to incrementally build a database of objects from one or multiple sessions.
    \item Object completion capabilities, facilitating \acronym{tsdf} merging of multiple occurrences of the same object.
    \item Integration of the proposed automatic object database building with a full scene reconstruction framework.
    \item The release of the Tango RGB-D dataset to serve as benchmarks for comparisons.
\end{itemize}

\section{Related Work}

In order to create a database of 3D objects, our approach first segments the input depth data, constructs a \acronym{gsm} and then inserts all the segments into the database.
Newly detected segments are then compared with the object models in the database to determine whether a new object model is required or an existing model needs to be updated.
In the following, we review literature relevant to the individual stages of our approach.

\textbf{Scene segmentation} is a very active research topic in the computer vision community and often a necessary prerequisite for certain tasks in robotics such as manipulation~\cite{kenney2009interactive}, object detection~\cite{guptaECCV14}, scene understanding \cite{karpathy2013object}, etc.
Numerous approaches have been developed in order to obtain meaningful segments from a scene. These can be extracted from RGB images~\cite{He2017MaskR}, RGB-D data~\cite{uckermann2013realtime, guptaECCV14, tateno2015real}, image sequences~\cite{shelhamer2016clockwork}, or 3D data such as point clouds~\cite{dube2018incremental}, meshes~\cite{felzenszwalb2004efficient, karpathy2013object}, voxel grids~\cite{Fehr2017}, or using deep neural networks~\cite{He2017MaskR, guptaECCV14, shelhamer2016clockwork}.
However, due to the requirement for large computational resources, deep neural networks are not applicable on most hardware constrained robotic systems.
%
%
Uckermann et al.~\cite{uckermann2013realtime} demonstrated a real-time capable segmentation method on depth images, based on a surface normal edge map which captures discontinuities of depth measurements, sudden changes of surface normals at object edges, and performs region growing of surface patches into segments.
The same idea was explored by Tateno et al.~\cite{tateno2015real} to provide a fast segmentation method for depth images.
Karpathy et. al~\cite{karpathy2013object} demonstrated an approach that partitions a scene mesh.

Our segmentation approach is similar to that of~\cite{uckermann20123d}.
We modified the normal estimation such that instead of just using three points in the vicinity of the midpoint, we use a kernel, resulting in a smoother normal map.
Additionally, instead of using the original edge map, we consider edges from a depth discontinuity map calculated based on~\cite{singh2014bigbird}.

Often, it is beneficial to fuse multiple segmented frames into a consistent map.
Finman et al.~\cite{finman2014efficient}  implemented a method that incrementally stores segmented depth data.
They use an incremental variation of the algorithm from~\cite{felzenszwalb2004efficient} to segment new data and a voting algorithm for recomputing parts of the \textbf{\acronym{gsm}} based on this new data, which ensures global consistency.
In their approach, the \acronym{gsm} relies on a 3D \acronym{tsdf} representation~\cite{curless1996volumetric}.
Tateno et al.~\cite{tateno20162} use a \acronym{tsdf} volumetric surface representation for the \acronym{gsm} and a SLAM system to keep track of the camera poses.
Our approach is based on~\cite{tateno20162}.
In contrast to~\cite{tateno20162}, we keep track of the entire history of segment labels associated to a \acronym{tsdf} voxel and assign that voxel to the segment with the highest count.
This approach increases robustness to noisy per-frame segmentation and allows us to deal more efficiently with merges of multiple segments into one, or splits of one segment into multiple ones.

From the \acronym{gsm}, raw segments can be extracted and added to the \textbf{database of objects}.
In contrast to some of the object datasets that are generated offline in controlled conditions~\cite{singh2014bigbird, salas2013slam++}, the \acronym{gsm} allows us to add segments to the database at any point during a mapping session.
Furthermore, if new information is observed, object models can easily be updated and completed online, without any post-processing required.
Dai et al. \cite{dai2017complete} developed a \textbf{shape completion} method that predicts and fills in the missing data from the input.
First, a fixed-size voxel volume is predicted using a 3D CNN and then a higher resolution model is synthesized based on an offline shape database.
A similar approach was developed by Han et al.~\cite{han2017high}.
However, both approaches can only complete shapes that they have been trained for.
Our approach does not require any a-priori knowledge about the objects and, furthermore, does not approximate objects but rather uses previously observed data to complete unobserved parts.
This also means that objects in the database will not be completed unless the missing data is observed in one of the sessions on at least one of the object instances.

Learning-based \textbf{object detection} methods usually try to classify patches in images~\cite{liu2016ssd,redmon2016yolo9000}, however, the number of classes such algorithms can detect is restricted by the training data provided to the algorithm.
Alternatively, database-assisted approaches such as~\cite{li2015database} rely on a limited number of pre-defined objects in the database, which are then detected in the scene using keypoints and descriptors.
In this work, we do not make assumptions on the objects in the scene.
Therefore, once we obtain the raw segments from the \acronym{gsm}, we use keypoints~\cite{Sipiran2011, zhong2009intrinsic} and descriptors~\cite{rusu2009fast} to represent those, and then match them to the previously obtained object models in the database.
Our approach is similar to~\cite{tateno2015real}, however, instead of global descriptors we use local descriptors and keypoints, allowing us to match individual parts of the models.
After we obtain the good correspondences from matching, we detect the pose using \acronym{ransac}~\cite{Fischler81CACM} and \acronym{icp}, as in~\cite{aldoma2012tutorial}.
The main difference of our approach to~\cite{tateno20162} is that we are able to, by merging individual segments, complete models of objects that we have not seen before, and therefore do not need to know them beforehand.

\section{Method}
In this section, we describe how we get from RGB-D images to object models in the database.
First, we use a geometric approach to \emph{segment} input depth images.
In a second step, we employ a \acronym{tsdf}-based \emph{\acronymfirst{gsm}}.
The \acronym{gsm} not only fuses depth measurements into a 3D reconstruction but obtains improved and temporally consistent labels, which ideally means one label per object instance.
Segments which remain unchanged in the \acronym{gsm} for a certain time are extracted (raw segments) and inserted into an initially empty \emph{database}.
Once these raw segments are in the database, we call them object models.
Afterwards, we match them to other models in an attempt to combine them into more complete and accurate object models.
Thus, we are trying to match them to either models of other instances of the same object or partial models of the same instance, created due to the oversegmentation of the \acronym{gsm}.
A third option is to match to object models from separate recordings.

\subsection{Depth Segmentation}
\label{sec:depth_segmentation}
\begin{figure}[!t]
\centering
\includegraphics[width=\columnwidth]{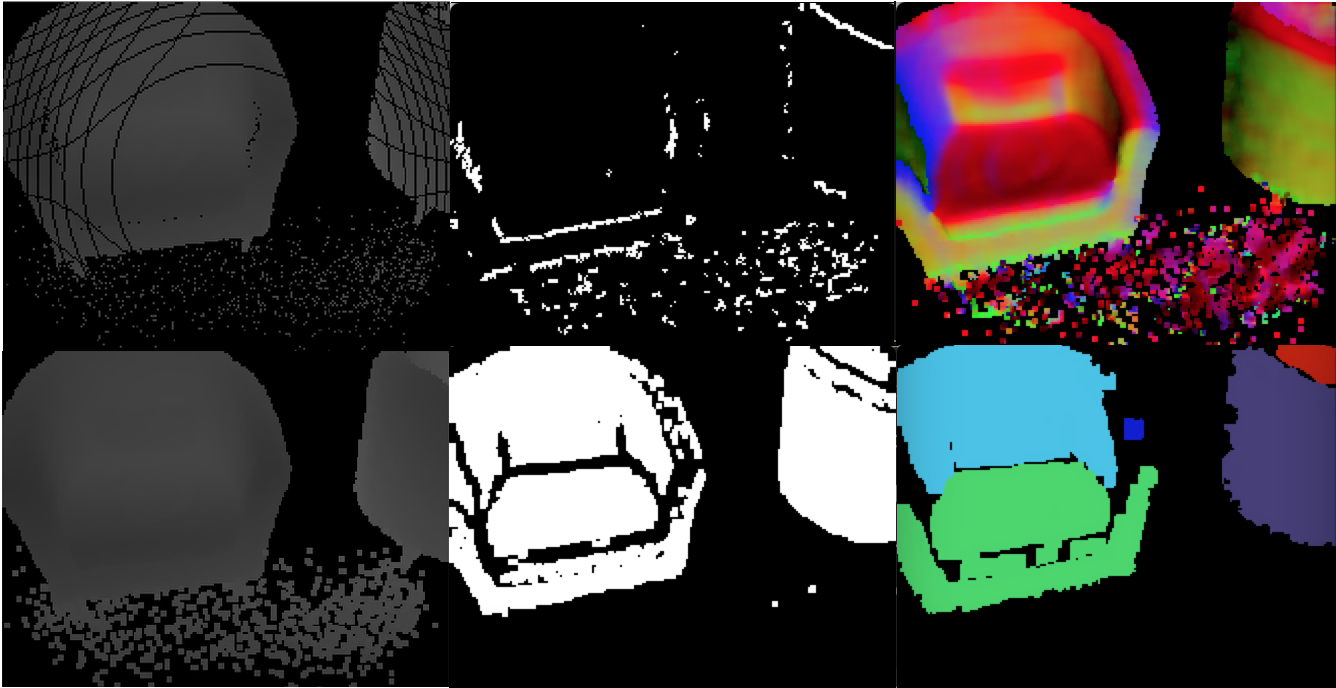}
\caption{The output of our depth segmentation for a single frame of the Tango dataset. Top left: rectified depth image, bottom left: filtered depth image, top center: depth discontinuity edges, bottom center: convexity map, top right: normal map, bottom right: labeled image.}
\label{fig:tango_ds}
\end{figure}
In order to segment RGB-D images, we use the fact that the objects we want to segment always have a closed shape.
Therefore, we apply similar filters as described in~\cite{uckermann2013realtime,tateno2015real} to the depth images.
In a first step, we inpaint the depth image to obtain more continuous areas with valid depth values.
This step is optional and its necessity depends on the specifications and configuration of the RGB-D sensor.
In a second step we detect edges that exhibit strong depth discontinuities.
Next, we compute surface normals based on a local pixel neighborhood and use them to determine the local convexity of each pixel.
In a final step, we combine the convexity map and the depth discontinuity filter to form closed regions and extract contours which we fill with a label.
However, these labels are not consistent across different frames and a nearly identical region might be assigned a different label in the next frame.
The depth measurements of each region can then be used to obtain a labeled 3D segment.
The results of the different steps are shown on a sample depth image of a Lenovo Tango Phab 2 Pro in Fig.~\ref{fig:tango_ds}

\subsection{Global Segmentation Map}
\label{sec:gsm}

The goal of the \acronym{gsm} is to merge the frame-wise segmentation into a more accurate and globally consistent object instance segmentation.
To that end, the segments extracted from each depth image are incrementally integrated into a \acronym{tsdf}-based voxel grid capable of storing and fusing labels for every voxel.
The proposed \acronym{gsm} serves the same purpose and shares some of the concepts with the surfel-based segmentation map proposed in~\cite{tateno2015real}.
Since our proposed system does not include a camera pose estimation stage, the camera poses for each depth map must be provided by an external estimation pipeline or ground truth data.

The \acronym{gsm} builds on top of Voxblox~\cite{oleynikova2017voxblox}, a real-time reconstruction framework based on a volumetric \acronym{tsdf} surface representation.
The Voxblox framework has been extended with a second volume, the label volume, storing the segment label associated with each voxel in the \acronym{tsdf} grid.
At each frame, the set of point clouds representing the 3D shapes of all the identified segments are fused into the voxel-based representation, with the \acronym{gsm} ensuring consistency in the segment labels across different frames.
The computational complexity of the method does not depend on the size of the map or the number of merged frames, and the resulting segmentation and reconstruction of the scene are obtained at interactive rates.

The main difference between our method and~\cite{tateno2015real} lies in the way the label volume is updated at each new segmented input depth map.
Instead of storing just one segment label and one confidence value at each voxel, we store the complete history of all the segment labels that have ever been merged into this voxel, together with the respective counts.
The label with the highest count is then set to be the main segment label associated with that voxel.
At the cost of additional memory usage, this approach is more robust towards noise in the per-frame segmentation.
Correct frame-wise segmentation outputs can contribute to the highest label observation count in a voxel, independently of whether they have been followed by a number of different improper ones.
This, in contrast, is not possible when approximating all the votes with a single value as in~\cite{tateno2015real}.
Furthermore, our method enables lossless merging of two or more segments in the map which have been detected to be part of the same object.
In the frames leading up to the segments getting merged, some of them are updated with the labels of the others, until it is detected that those multiple different labels are actually one.
When only one label count is stored in a voxel as in~\cite{tateno2015real}, these updates lead to the lowering of the previously gained confidence for the merged segments, while our approach allows to recover the total number of correct votes for each voxel by simply summing up the label counts for the merged segments.

For each 3D segment representing a distinct object in the scene and uniquely identified by a label, the \acronym{gsm} also implicitly stores its pose in a world frame.
A segment is extracted from the \acronym{gsm} when none of the voxels corresponding to the segment have been updated for a certain period of time.
The \acronym{tsdf} grid of this extracted raw segment is sent to the database together with the corresponding segment label and its world pose.

\subsection{Incremental Object Database}
\label{sec:incremental_object_database}
\begin{figure}[!t]
\centering
\begin{tabular}{cc}
\subfloat[Erroneous match but good \acronym{icp} score, \acronym{tsdf} verification required ]{\includegraphics[width=0.45\columnwidth]{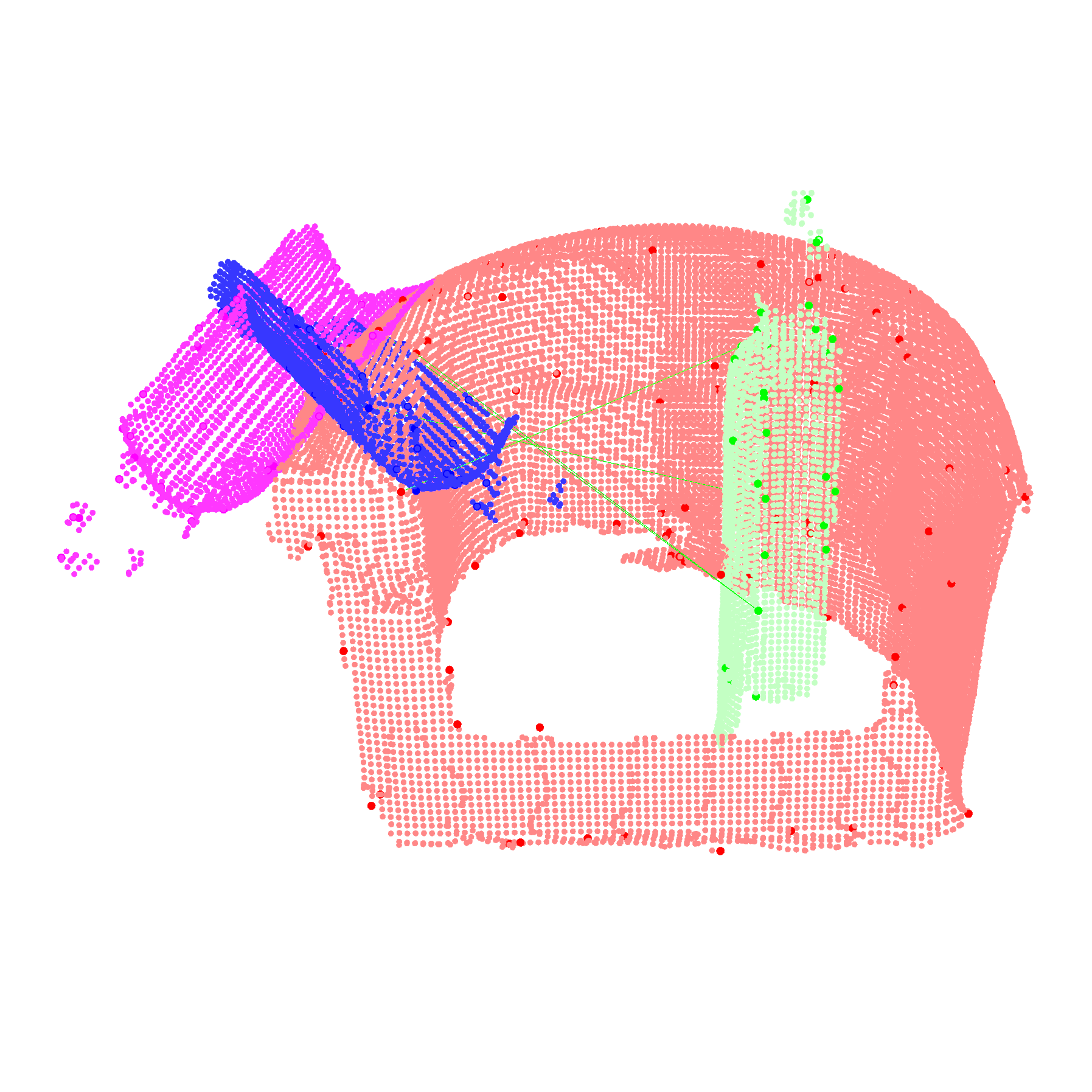}}&
\subfloat[Correct match with good \acronym{icp} score]{\includegraphics[width=0.45\columnwidth]{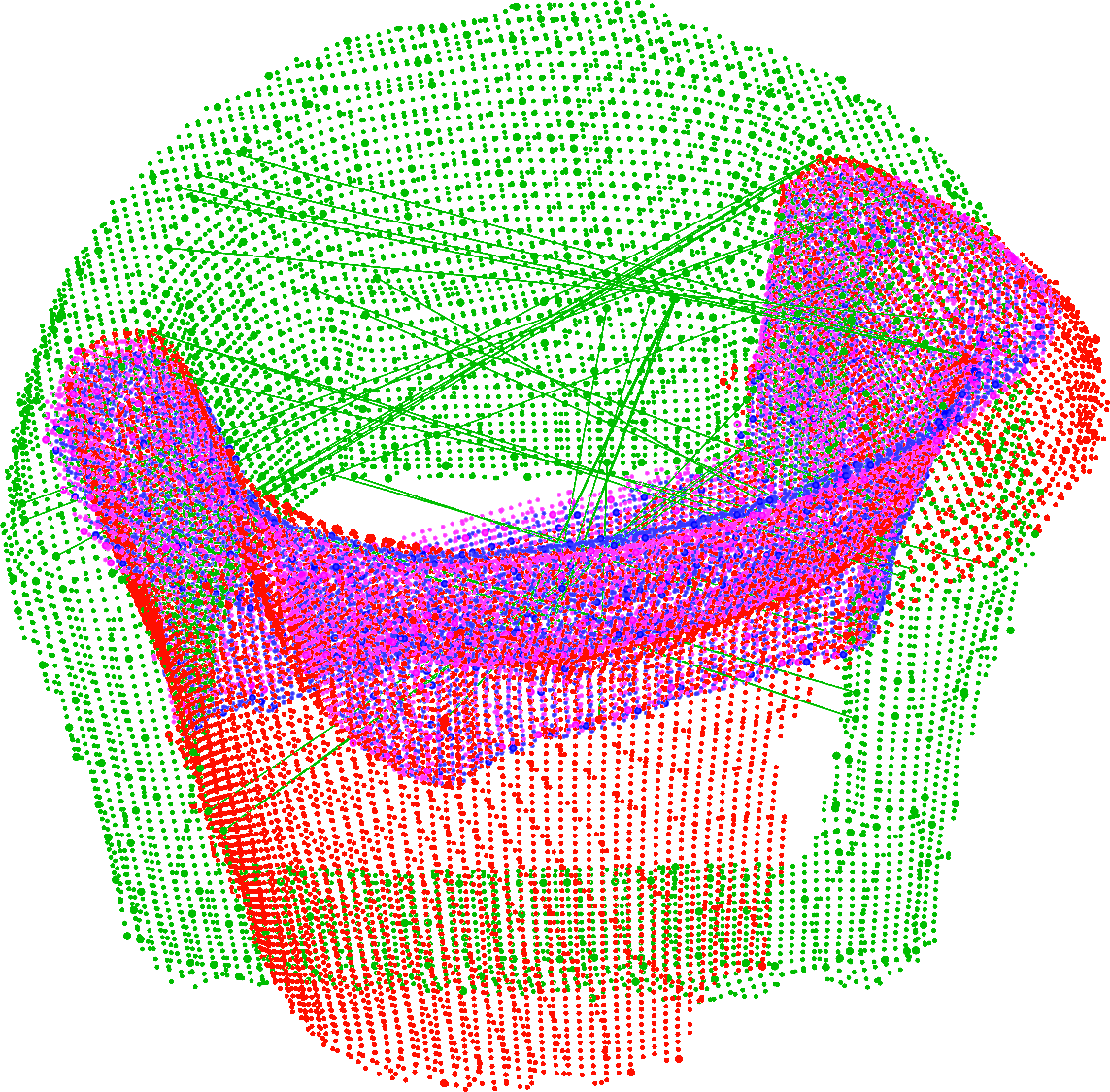}}
\end{tabular}
\caption{The keypoints are matched among two point clouds (green and red), the green cloud is registered to the red one using \acronym{ransac} (blue) and this registration is refined using \acronym{icp} (magenta). Green lines indicate the good matches. Both images result in a match with a good \acronym{icp} \acronym{rmse}. (a) will be rejected by the geometric verification step described in Section~\ref{sec:merging}, whereas (b) is a correct match.}
\label{fig:matching}
\end{figure}
The object models $s_i$ in the database $\Omega$ consist of the following components:
a set $\mathbb{T}_i$ of poses $\boldsymbol{T}_k$\footnote{All the transformations with one subscript denote transformations from the world origin to the object model base frame.} for all model observations constituting $s_i$, where the first element of the set $\mathbb{T}_i$ defines the base frame of the model $s_i$, a \acronym{tsdf}-grid $t_i$ retaining the 3D shape of the model, the corresponding surface as a point cloud with normals $\boldsymbol{c}_i$, as well as 3D keypoints $\boldsymbol{k}_i$, and their 3D feature descriptors $\boldsymbol{d}_i$.

Upon insertion into the database, the labeled segments extracted from \acronym{gsm} are not full object models yet.
In this section we first explain how we complete the object model with keypoints and descriptors followed by the matching and merging of object models.

\subsubsection{Point Cloud Extraction}
\label{sec:point_cloud_extraction}
In order to merge raw segments extracted from the \acronym{gsm} into complete object models, we need to match and register these segments.
First we use the marching-cubes surface reconstruction algorithm~\cite{lorensen1987marching} on the \acronym{tsdf} grid to obtain a point cloud with surface normals, which is required to extract keypoints and descriptors for matching.
We use a \acronym{ransac}-based planarity check to exclude planar segments from merging as they do not provide enough constraints to allow for meaningful matches, based on their geometry.

\subsubsection{Keypoints and Descriptors}
\label{sec:keypoints_and_descriptors}
From the remaining segments, we extract keypoints and describe their neighborhood in the point cloud using descriptors.
We use a combination of \acronym{iss}~\cite{zhong2009intrinsic} and Harris3D~\cite{Sipiran2011} keypoints.
The two types of keypoint detectors complement each other well for our application.
The \acronym{iss} detector is very efficient at the cost of reduced repeatability~\cite{Tombari2013performance}, therefore, we used it to extract a larger number of keypoints even in smoother areas, whereas Harris3D was tuned to provide fewer but more repeatable keypoints.

To describe the resulting keypoints, we use \acronym{fpfh} descriptors~\cite{rusu2009fast}, a very efficient but low-dimensional descriptor that is based on the surface normals in a spherical neighborhood of radius $r$.
While the efficiency of \acronym{fpfh} allows fast matching to a large number of object model candidates, it requires strong geometric consistency checks to compensate for the limited expressiveness.

\subsubsection{Matching and Registration}
\label{sec:matching_and_registration}
\begin{figure}[!t]
\begin{tabular}{cc}
\subfloat[Match rejected by model to model verification:
The surfaces are registered upside down, thus the
\acronym{icp} score is good, but the \acronym{tsdf} values have opposite signs.
]{\includegraphics[width=0.45\columnwidth]{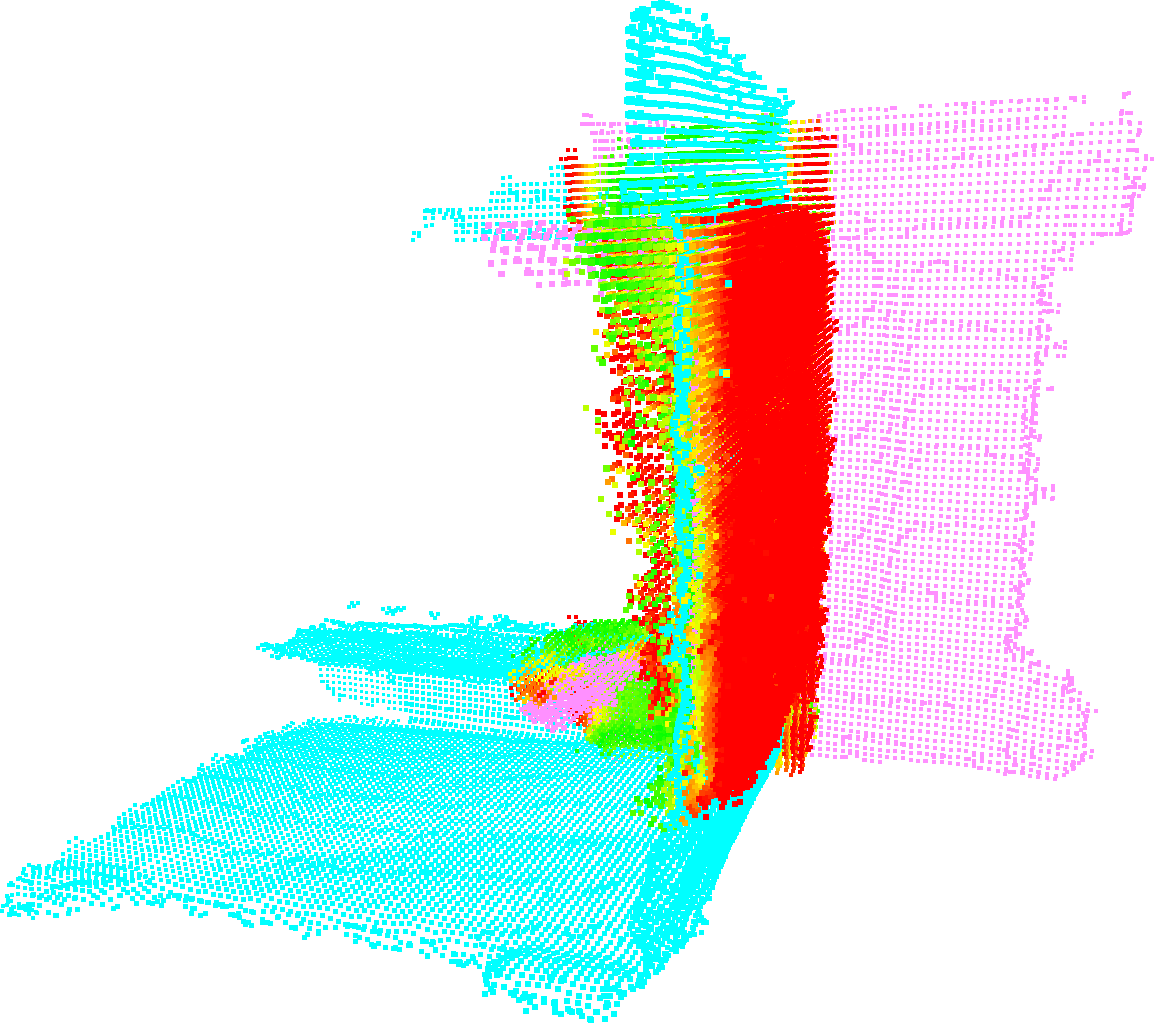}} &
\subfloat[Model to model verification failed to reject match between a chair and a fire extinguisher. Model to scene verification required.]{\includegraphics[width=0.45\columnwidth]{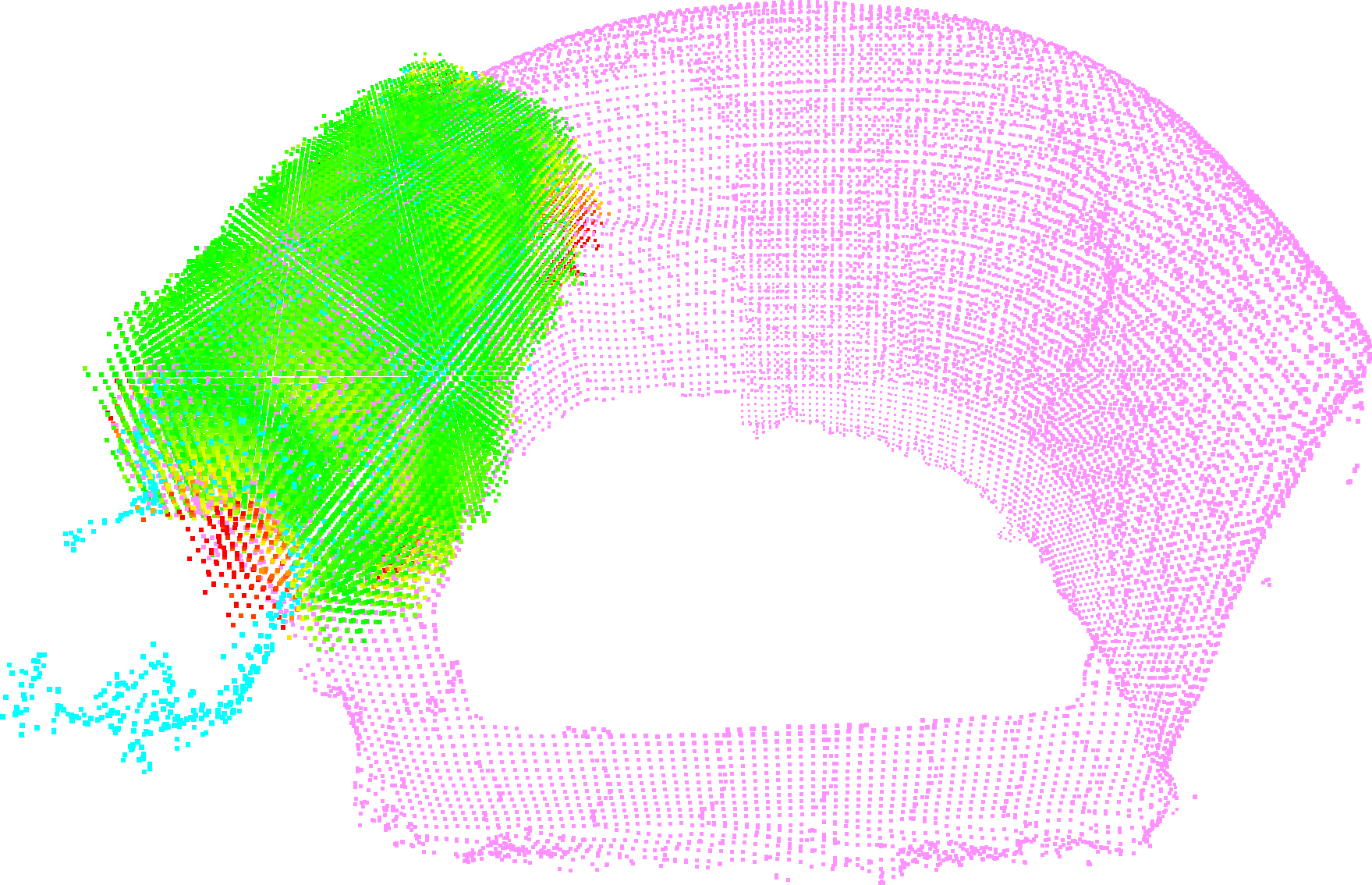}} \\
\subfloat[Match accepted by model to model verification.]{\includegraphics[width=0.45\columnwidth]{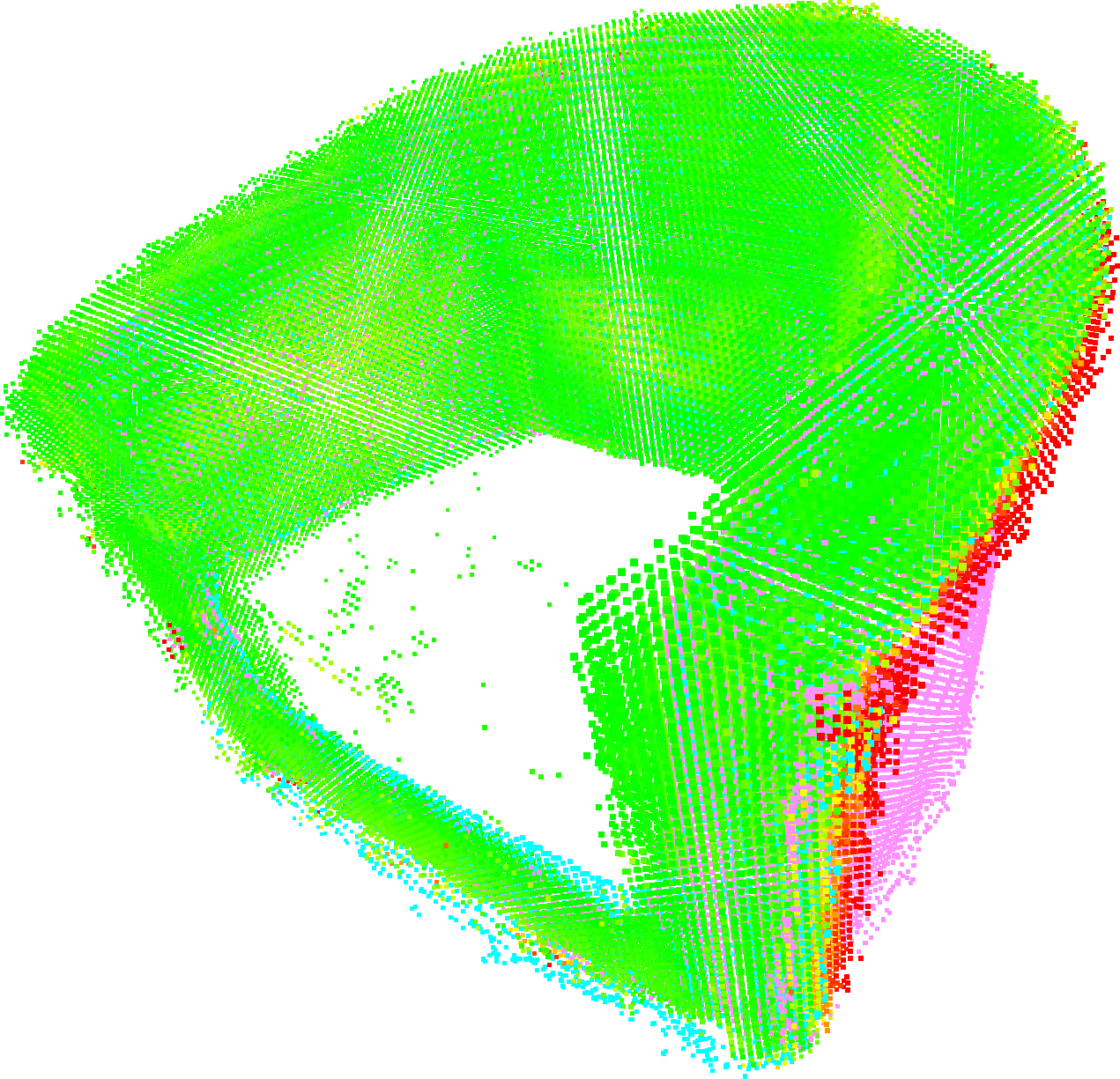}} &
\subfloat[Model to scene verification. Correctly rejects erroneous match of (b) and accepts match of (c).]{\includegraphics[width=0.45\columnwidth]{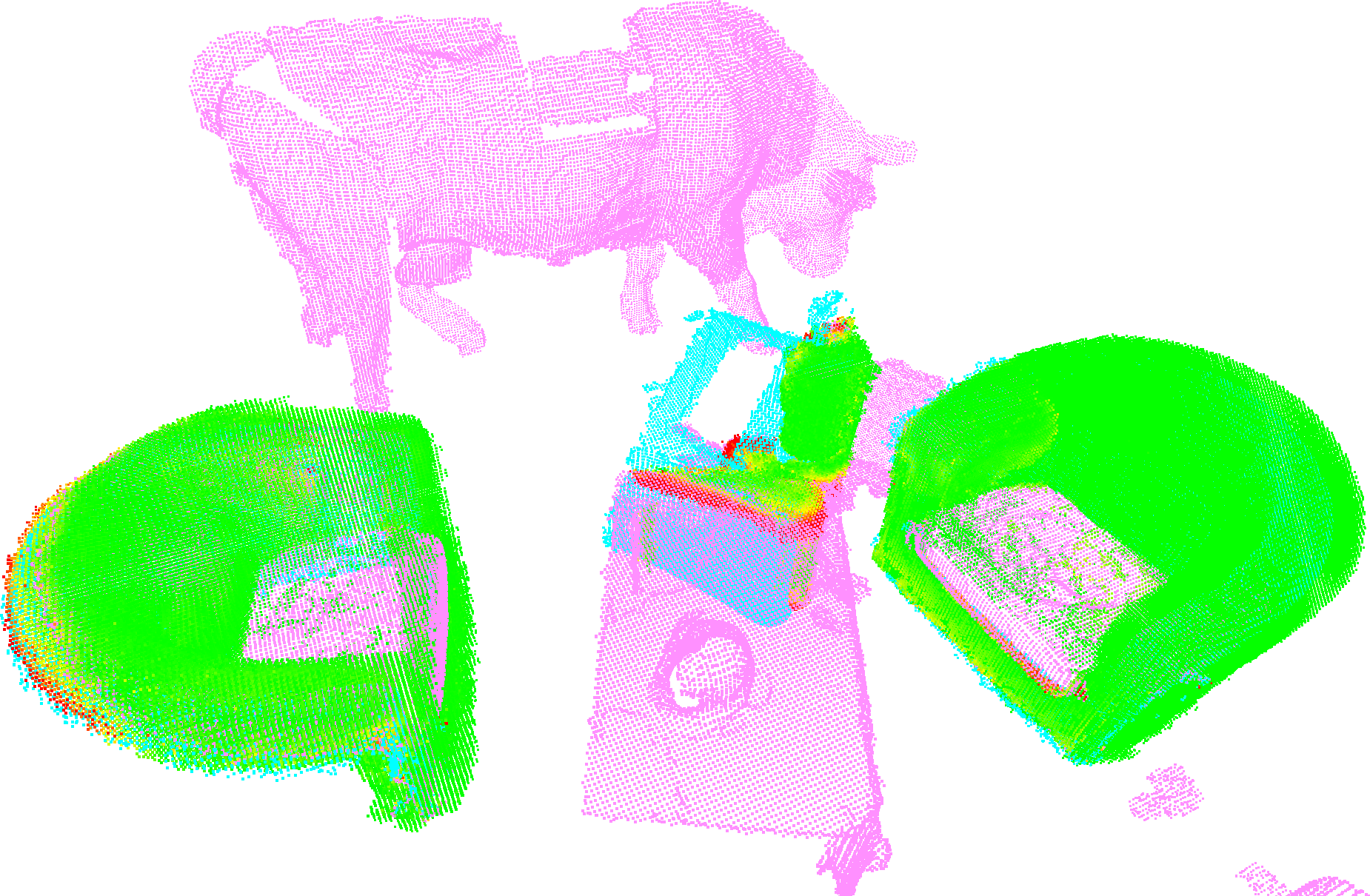}}
\end{tabular}
\caption{Geometric verification of object models.
The evaluated models are represented by the \textit{cyan} and \textit{magenta} point clouds.
The \acronym{tsdf} errors between the two voxel grids of the models are visualized as one point per overlapping voxel, colored \textit{green} to \textit{red}, where \textit{red} represents a high error.}
\label{fig:verification}
\end{figure}
After obtaining an object model as defined at the beginning of this section, we want to find database entries that match this newly created object model.
In contrast to other state-of-the-art approaches~\cite{mattausch2014object,salas2013slam++}, we do not assume any prior information about the environment, local relation between several segments, their semantics, or rely on any prior object models.
However, it is possible to incorporate prior knowledge (e.g. semantic segmentation) to support the matching process and improve speed and accuracy.

We apply a three-step registration process that allows for global registration of two point clouds based on their geometry.
First, we conduct an efficient nearest neighbor search in descriptor space using a kd-tree to obtain matching 3D descriptors.
For every keypoint we select the best $k_{NN}$ matches with a descriptor similarity score above $t_{sim}$.

Secondly, we use \acronym{ransac} to find a geometrically consistent set of inlier matches as well as the initial transformation $\boldsymbol{T}_{i,j}^{coarse}$ between models $s_i$ and $s_j$.
If no consistent set of feature matches is found, the candidate is rejected.

In the refinement step, Point-to-Plane \acronym{icp} is applied to the coarsely registered models, yielding an improved registration transformation $\boldsymbol{T}_{i,j}^{fine}$.
We apply a threshold $e_{ICP}^*$ to the \acronym{rmse} of \acronym{icp} to reject unsuccessful registrations.
The final transformation between the models $s_i$ and $s_j$ is then obtained as follows:
\begin{equation}
\boldsymbol{T}_{i,j}=\boldsymbol{T}_{i,j}^{fine}\boldsymbol{T}_{i,j}^{coarse} \enspace .
\end{equation}
Fig.~\ref{fig:matching} shows two examples of the descriptor matching and the intermediate states of the model registration process.

\subsubsection{Merging and Verification}
\label{sec:merging}
After obtaining a registration for an object model candidate, we verify the match using a two-stage geometric consistency check, only if it succeeds, we merge model $s_i$ into model $s_j$.
We perform these verification steps based on the models' \acronym{tsdf} grids $t_i$ and $t_j$.
First, we transform the \acronym{tsdf} grid $t_i$ of model $s_i$ into the \acronym{tsdf} grid $t_j$ of model $s_j$ using trilinear interpolation
\begin{equation}
t'_i = \Xi(\boldsymbol{T}_{j,i}\cdot t_i) \enspace.
\end{equation}

The first model to model verification step is performed by counting the overlapping voxels ($o_{TSDF}$) and calculating the \acronym{rmse} of all voxel pairs ($e_{TSDF}$) of the aligned voxel grids.
We enforce a maximum \acronym{rmse} ($e_{TSDF}^*$) and a minimum overlap ($o_{TSDF}^*$) between $t'_i$ and $t_j$:
\begin{equation}
e_{TSDF}^*>\sqrt{(t_j - t'_i)^2} \enspace ,
\end{equation}
\begin{equation}
o_{TSDF}^* < \left|t_j \cap t'_i\right|
\enspace.
\end{equation}
If the match is accepted, the aligned grids are merged by taking the weighted average of each voxel pair:
\begin{equation}
\hat{t}_j = t_j \oplus t'_i
\enspace.
\end{equation}
In a final model to scene verification step, the merged \acronym{tsdf} grid $\hat{t}_j,$ is transformed into all observed locations in the \acronym{gsm} and their geometric consistency with the scene \acronym{tsdf} $t_{scene}$ is verified based on the same metrics as above.
Both verification steps are depicted in Fig.~\ref{fig:verification}.
The whole process of inserting a new segment into the database is furthermore outlined in Algorithm~\ref{object_completion}.
\begin{algorithm}
\scriptsize
\caption{Incremental 3D object database}
\label{object_completion}
\begin{algorithmic}[1]
\Procedure{\textnormal{match\_and\_register()}}{}
\State $\boldsymbol{T}_{i,j}^{coarse}\gets $matching\_and\_RANSAC$(\boldsymbol{k}_i, \boldsymbol{d}_j, \boldsymbol{k}_j, \boldsymbol{d}_j)$
\State $\boldsymbol{T}_{i,j},e_{ICP}\gets $ICP$(\boldsymbol{c}_j,\boldsymbol{T}_{i,j}^{coarse}\boldsymbol{c}_i)$
\EndProcedure
\Procedure{\textnormal{insert\_segment\_in\_database}($s_i$)}{}
    %
        %
        \State candidates $ \gets \emptyset$
        \While{$s_j\gets$load\_next\_segment$(\Omega)$}
            \State  $\boldsymbol{T}_{i,j},e_{ICP} \gets $match\_and\_register$(s_i,s_j)$
            \If{$e_{ICP} > e_{ICP}^*$}
                \State \textbf{skip} $s_j$
            \EndIf
            \State $t'_i \gets \Xi(\boldsymbol{T}_{j,i}\cdot t_i)$
            \State $e_{TSDF},o_{TSDF}  \gets $geom\_consistency$(t'_i,t_j)$
            \If{$e_{TSDF} > e_{TSDF}^* \lor o_{TSDF} < o_{TSDF}^*$}
               \State \textbf{skip} $s_j$
            \EndIf
            \State $\hat{t}_j \gets t_j \oplus t'_i$
            \ForAll{$\boldsymbol{T}_k \in \mathbb{T}_i \cup \mathbb{T}_j$}
                \State $\hat{t}'_j \gets \Xi(\boldsymbol{T}_k\cdot \hat{t}_j)$
                \State $e_{TSDF},o_{TSDF} \gets $geom\_consistency$(\hat{t}'_j, t_{scene})$
                \If{$e_{TSDF} > e_{TSDF}^* \lor o_{TSDF} < o_{TSDF}^*$}
                    \State \textbf{skip} $s_j$
                \EndIf
            \EndFor
            \State $\hat{s}_j\gets $compute\_object\_model$(\hat{t}_j)$
            \State candidates $ \gets (\hat{s}_j,e_{TSDF})$
        \EndWhile
        \State $\hat{s}\gets $get\_lowest\_rmse$($candidates$)$

        \State $\Omega \gets $insert$(\hat{s})$
        %
    %
\EndProcedure
\end{algorithmic}
\end{algorithm}

\section{Experiments}
\begin{figure}[!t]
\centering
\includegraphics[width=\columnwidth]{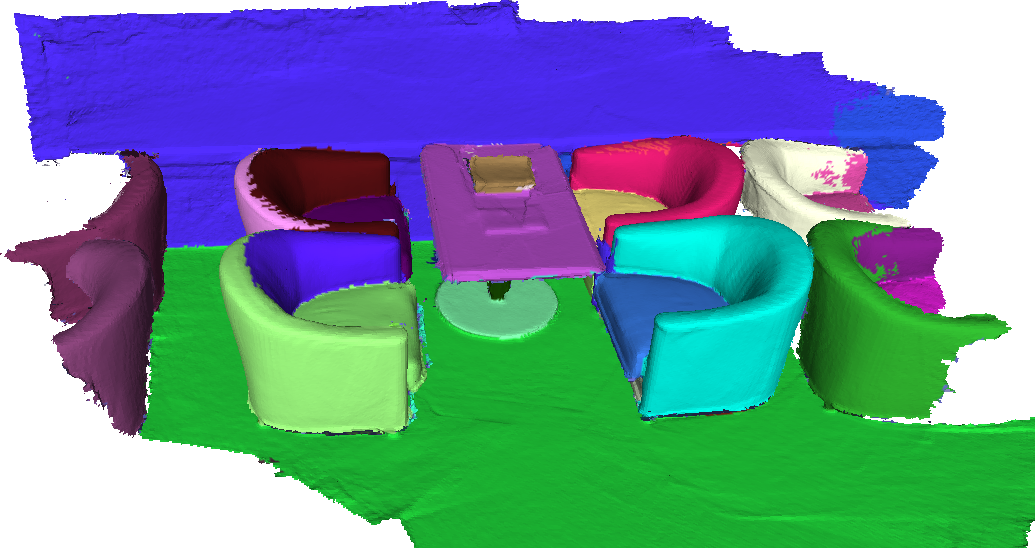}

\vspace{-0.5cm}

\includegraphics[width=\columnwidth]{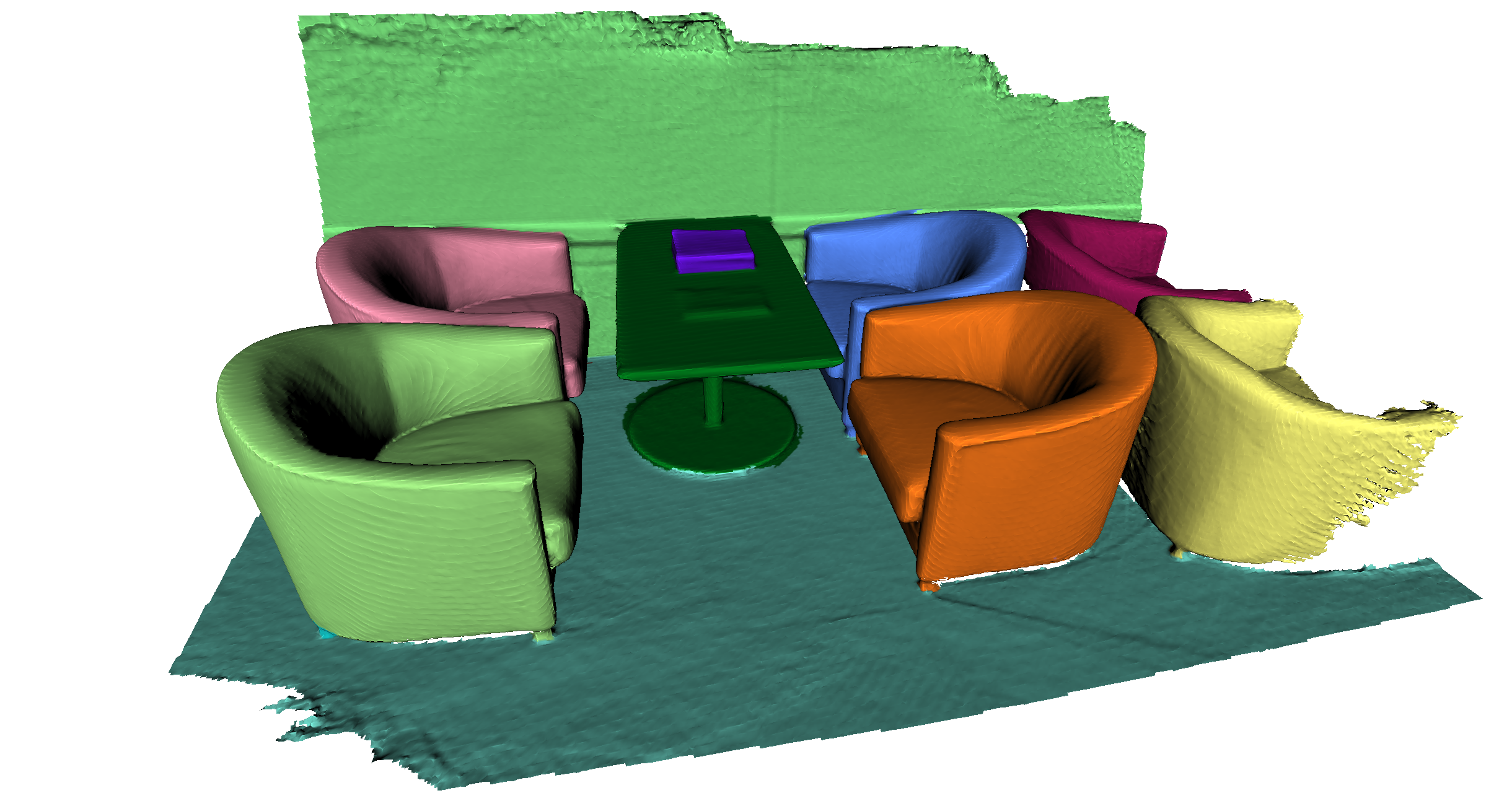}
\caption{\textit{top:} The output of \acronym{gsm} on the SceneNN~\cite{scenenn-3dv16}, scene 066. Different colors indicate different segments. \textit{bottom:} Ground truth segmentation reported by the authors of the dataset.}
\label{fig:scenenn_066_gsm}
\end{figure}
\begin{figure}[!t]
\centering
\includegraphics[width=\columnwidth]{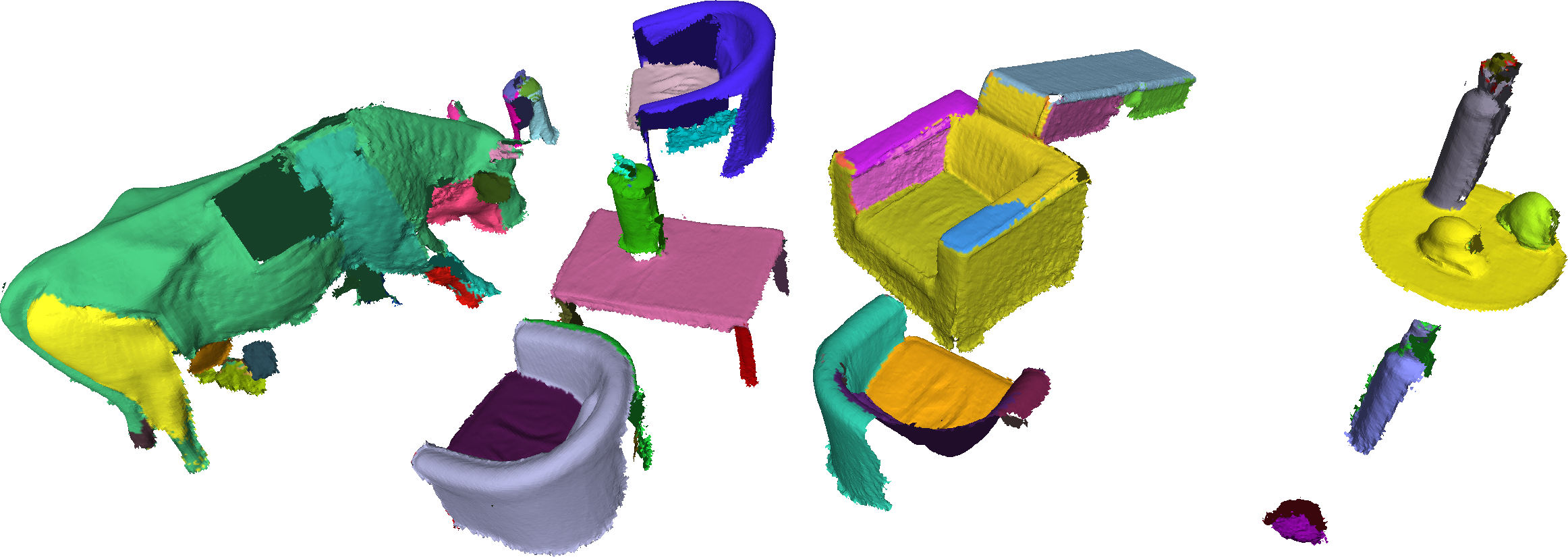}

\vspace{0.5cm}

\includegraphics[width=\columnwidth]{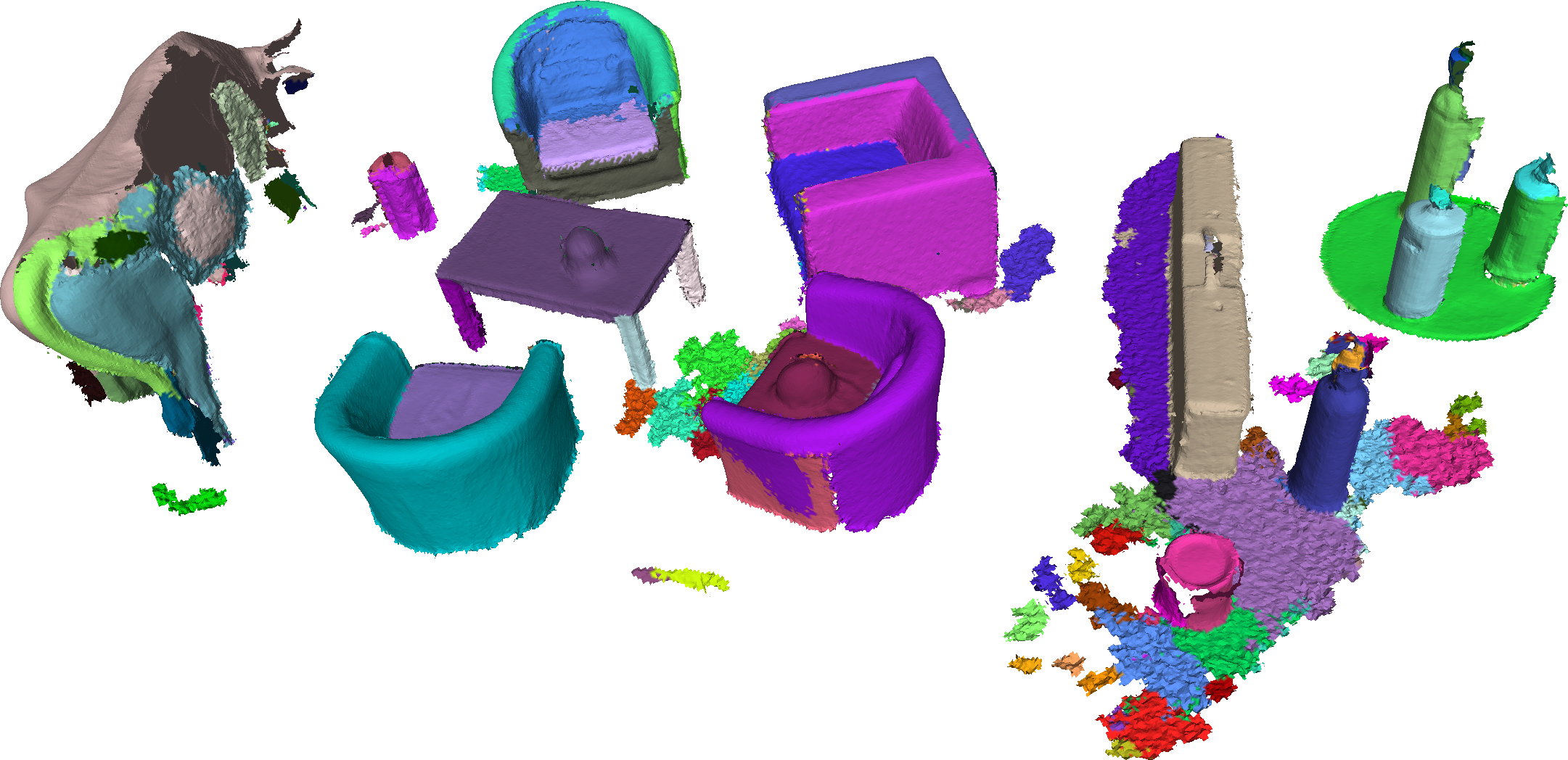}

\vspace{0.25cm}

\includegraphics[width=\columnwidth]{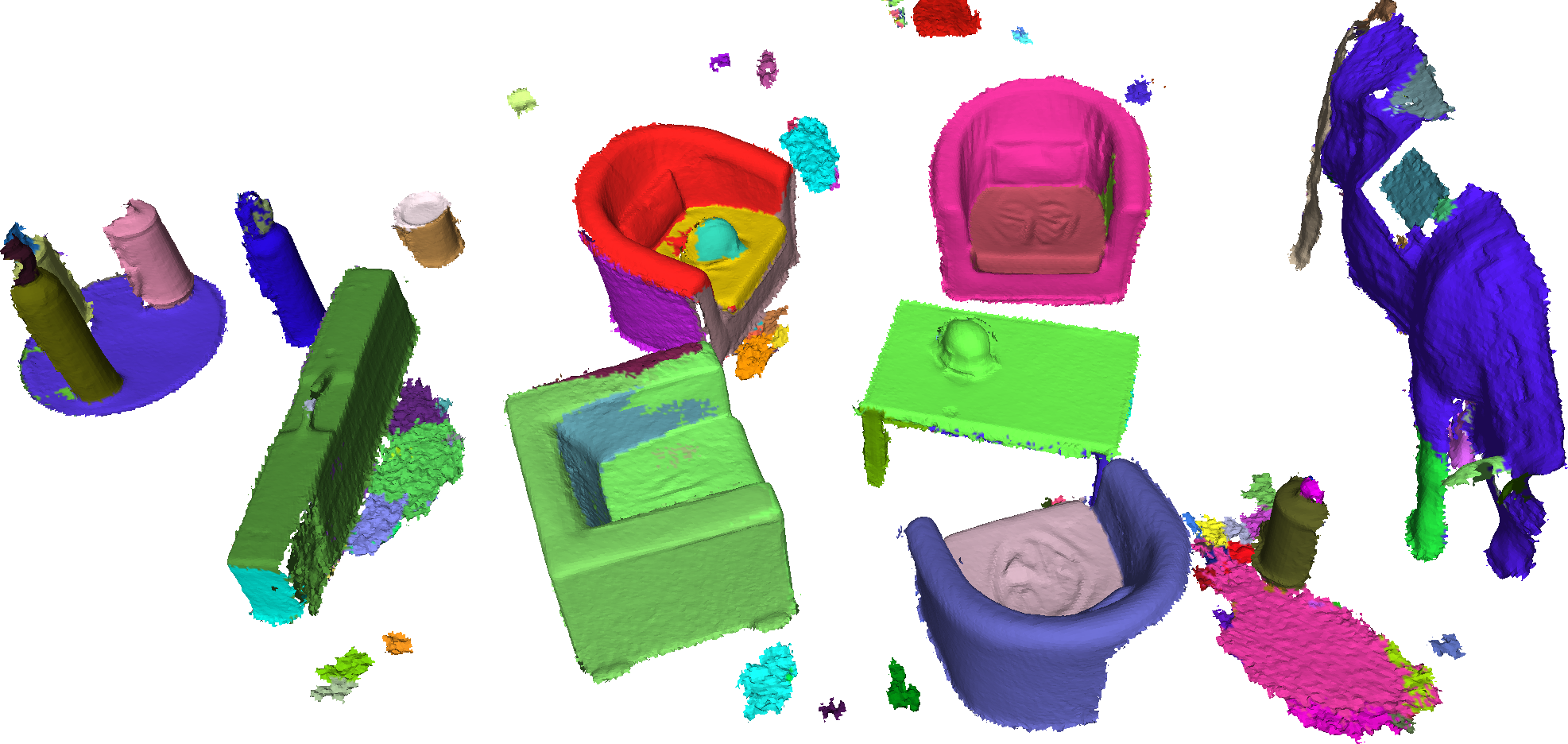}
\caption{Labeled surface mesh of the \acronym{gsm} for the three large Google Tango dataset scenes. Every scene contains eleven distinguishable objects, some of which appear multiple times, e.g. the round chair.}
\label{fig:tango_gsm}
\end{figure}

To evaluate our system, we perform experiments on sequence 66 of the SceneNN dataset~\cite{scenenn-3dv16}, where multiple identical objects are present, shown in Fig.~\ref{fig:scenenn_066_gsm}, and on four indoor datasets collected with a Tango phone, and released with this paper\footnote{The datasets are available at \href{https://projects.asl.ethz.ch/datasets/doku.php?id=iros2018incrementalobjectdatabase}{https://projects.asl.ethz.ch/datasets/}.}, see Fig.~\ref{fig:tango_gsm}.

The indoor datasets were recorded with a Lenovo Phab 2 Pro, which is equipped with the Google Tango sensor suite, i.e., an RGB-D sensor and a grayscale fisheye camera for motion tracking. The Google Tango system provides accurate camera pose estimates using a keyframe-based visual inertial SLAM algorithm, including post-processing with loop closure detection and full bundle-adjustment.
The datasets were recorded in a lounge area, with different arrangements of objects, i.e., 17 objects from 11 different object categories~(\emph{cow~(1), round chair~(3), cube chair~(1), fire extinguisher~(3), gas cylinder~(2), helmet~(2), round table~(1), square table~(1), large box~(1), small container~(1), floor~(1)}).

Firstly, we show the segmented reconstruction output of the \acronym{gsm}.
Secondly, we evaluate the benefits of an incremental object database with its merging capabilities.
Finally, we demonstrate the system for single- and multi-session scene completion.
We report the most relevant parameters of the system in Table~\ref{tab:parametrization}.

\subsection{Depth Segmentation and \acronym{gsm}}
The depth segmentation of a single depth image on the Tango dataset is shown in Fig.~\ref{fig:tango_ds}.
After the integration of multiple segmented frames into the \acronym{gsm}, we obtain the initial segmented scene, depicted in Fig.~\ref{fig:tango_gsm}.
In this scene, multiple instances of the same object can be observed, e.g., multiple chairs of the same kind.
From the \acronym{gsm} output, it is visible that some objects are segmented into multiple parts, showing the need to merge segments and observation of repetitive objects.
Similarly, the SceneNN dataset contains multiple instances of one chair and, hence, akin observations can be made on the \acronym{gsm} output, shown in Fig.~\ref{fig:scenenn_066_gsm}.

The implementation of depth segmentation is currently the bottleneck of the \acronym{gsm} pipeline in terms of processing speed, able to process VGA resolution depth maps at only $\sim3$\,Hz.
The \acronym{gsm} can integrate segmented depth maps at VGA resolution at $\sim8$\,Hz.
Hence, for real-time performance, VGA depth maps need to be down-sampled.
The Tango devices, on the other hand, provide depth maps at a lower resolution (224x172\,px) at 5\,Hz and can be processed in real-time.
\begin{table}[!t]
\scriptsize
\centering
\caption{Parameters of \acronym{iss}/Harris3D keypoint detectors, FPFH descriptor, Kd-tree Matching, and \acronym{tsdf} merging.}
\label{tab:parametrization}
\begin{tabular}{@{}p{1.65cm}@{\hspace{0.15cm}}p{1.5cm}@{\hspace{0.15cm}}p{1.2cm}@{\hspace{0.15cm}}p{1.6cm}@{\hspace{0.15cm}}p{2.05cm}@{}}\toprule
Harris3D & ISS & FPFH & Matching & Merging\\ \midrule
$r_H=\SI{3}{cm}$ & $r_{b}=\SI{2}{cm}$ & $r=\SI{10}{cm}$ & $t_{sim}=0.4$ & $e_{TSDF}^*=\SI{2}{cm}$\\
$k_{NN}^{min}=10$ & $r_{s}=\SI{6}{cm}$ &  & $k_{NN}=5$ &  $o_{TSDF}^*=10^4$\\
$t_{curv}=0.07$ & $r_{nm}=\SI{8}{cm}$ &  & $e_{ICP}^*=\SI{2}{cm}$  &\\
$t_{resp}=10^{-6}$ & $k_{NN}^{min}=5$ &  &  &\\
 \bottomrule
\end{tabular}
\end{table}

\subsection{Incremental Object Database}
\begin{figure}[!t]
\centering
\includegraphics[width=\columnwidth]{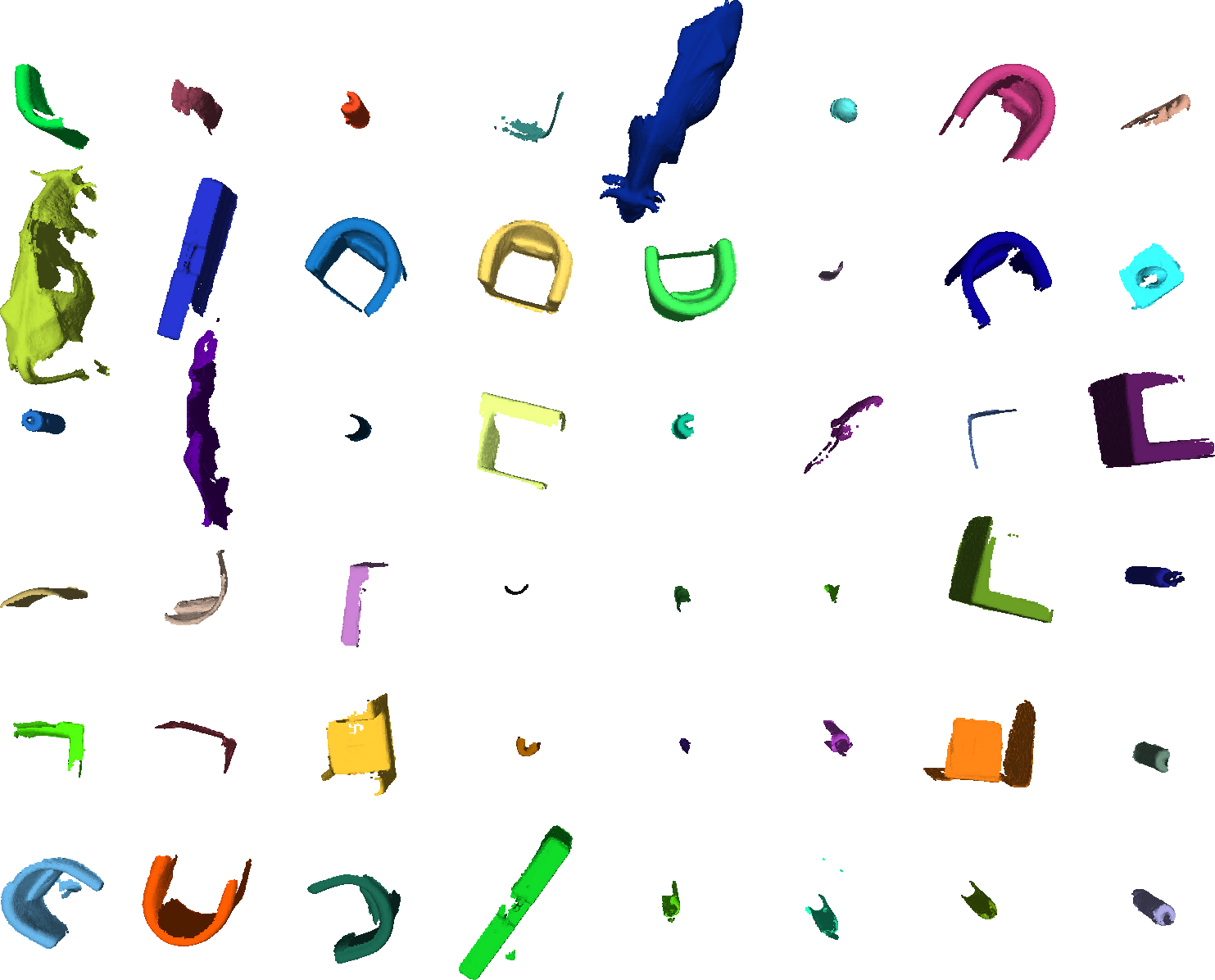}

\vspace{0.5cm}

\includegraphics[width=\columnwidth]{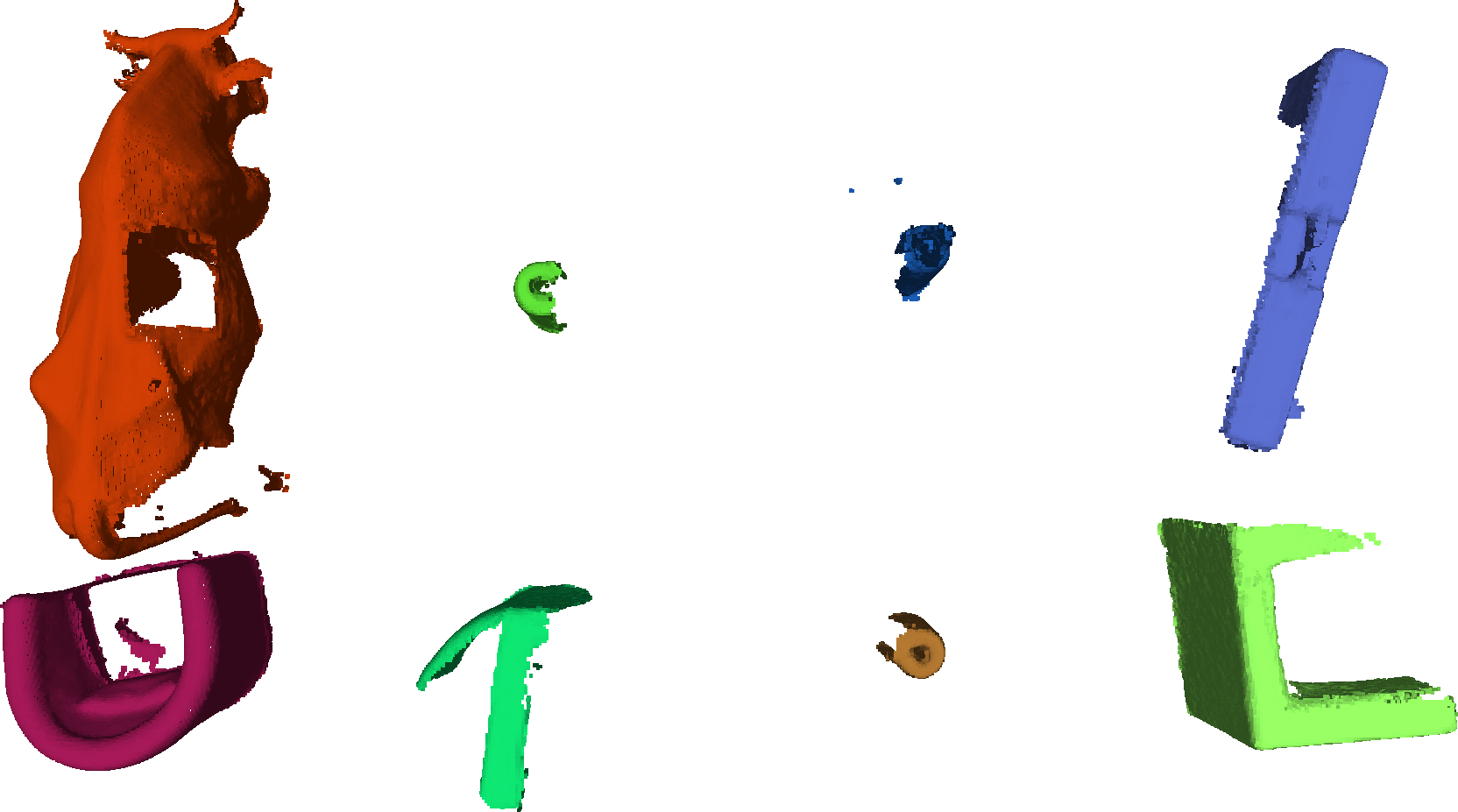}
\caption{Extracted \acronym{gsm} raw segments from three Google Tango recordings of a scene that contained eleven distinguishable objects, with one or more instances of each object.
From originally 330 raw segments, 48 mergeable (non-planar) raw segments remained (\textit{top}).
They are processed by the database and combined into 27 object models, of which 8 (\textit{bottom}) are the result of merging several raw segments:
cow (3), round chair (10), fire-extinguisher (4), gas cylinder (4), box(2), cube chair (2).
The algorithm produces only a single erroneously merged object model (\textit{2nd object bottom row}).
}
\label{fig:tango_merged_and_raw_mergeable_multi_session}
\end{figure}

The main goal of our incremental object database is to improve the quality and completeness of our object models by merging the knowledge of multiple observed object instances.
We show this and the reduction of raw segments, extracted from the \acronym{gsm}, by matching and merging them to object models previously inserted into the database if the verification steps indicate equal object instances, in Fig.~\ref{fig:tango_merged_and_raw_mergeable_multi_session}.

From an initial set of 330 raw segments, recorded over three sessions, we recognize 48 raw mergeable segments that are reduced to 27 database object models.
Six of these objects consist of two or more merged segments.

In Table~\ref{tab:computation_times}, we report the computational times of the individual steps of the database from point cloud extraction from \acronym{tsdf} grids up to the verification steps among segments and to the scene, computed on a Intel Xeon CPU~@~2.80\,GHz (8 cores) running in a single thread.
On average we spend \SI{32.0}{ms} on extracting the point cloud, \SI{95.4}{ms} on keypoint extraction and \SI{107.3}{ms} on computing descriptors for a single segment.
The timings show that the most time consuming step is the object model matching, more specifically the \acronym{icp} refinement step.
Please note that the timings reported for matching, merging and verification are highly dependent on the dataset and the resulting database.
\begin{table}[!t]
\scriptsize
\centering
\caption{Computation times for processing the large three scenes of the Tango dataset: Point cloud extraction (\textbf{P},~\ref{sec:point_cloud_extraction}) from the \acronym{tsdf} grid, the Keypoint extraction (\textbf{K},~\ref{sec:keypoints_and_descriptors}), the Descriptor computation (\textbf{D},~\ref{sec:keypoints_and_descriptors}), the Matching \& Registration (\textbf{MR},~\ref{sec:matching_and_registration}), the Merging \& model to model Verification (\textbf{MV},~\ref{sec:merging}), and model to Scene Verification (\textbf{SV},~\ref{sec:merging}).}
\label{tab:computation_times}
\begin{tabular}{@{}cccccccc@{}}\toprule
         & \textbf{P} & \textbf{K} & \textbf{D} & \textbf{MR} & \textbf{MV} & \textbf{SV} & $\boldsymbol{\Sigma}$\\ \midrule
Time [s] & 11.9 & 22.4 & 24.4 & 1267.6 & 21.6 & 24.7 & 1372.6 \\
\bottomrule
\end{tabular}
\end{table}

\subsection{Single- \& Multi-Session Scene Completion}
\begin{figure}[!t]
\centering
\includegraphics[width=\columnwidth]{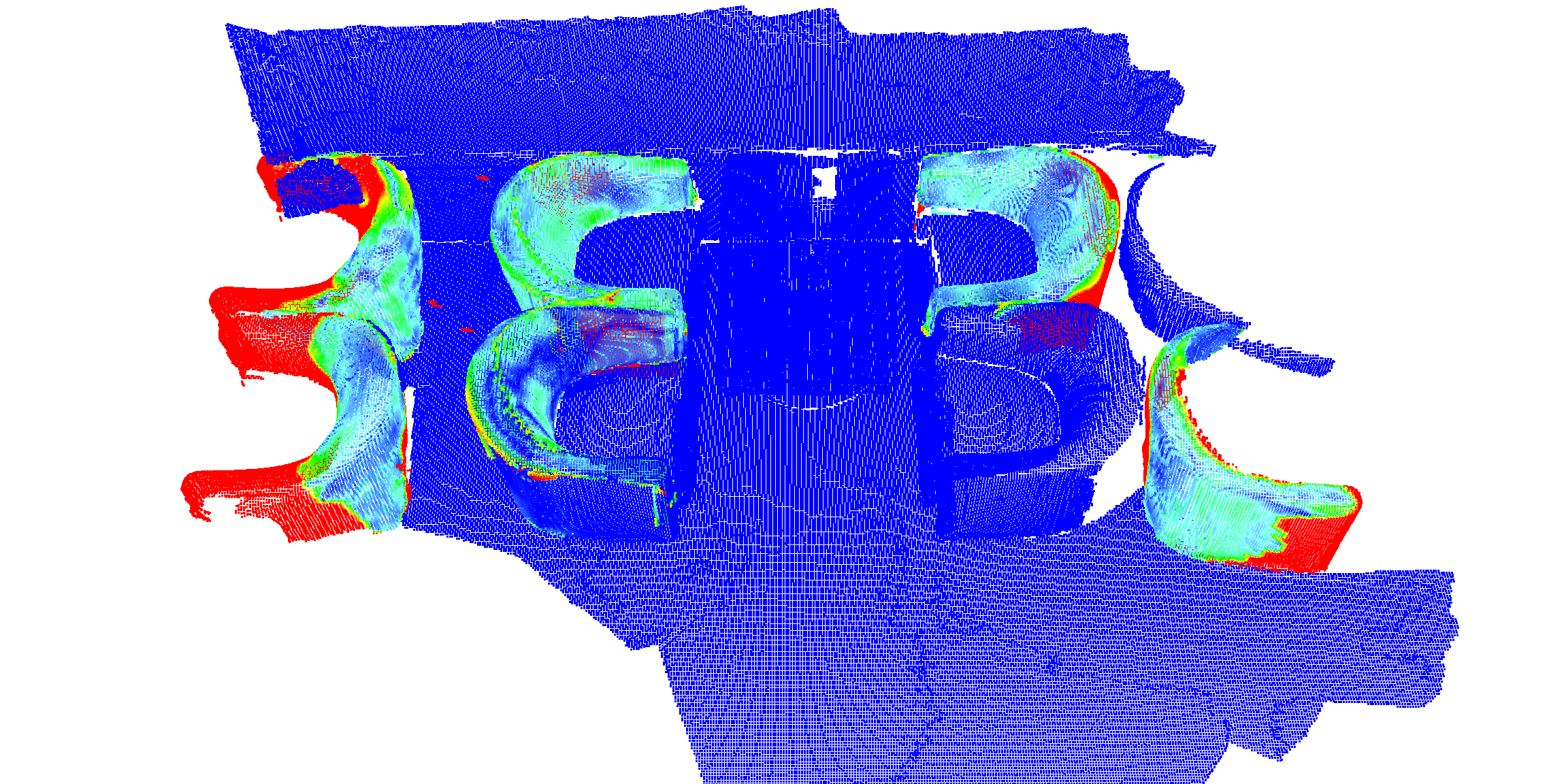}
\includegraphics[width=\columnwidth]{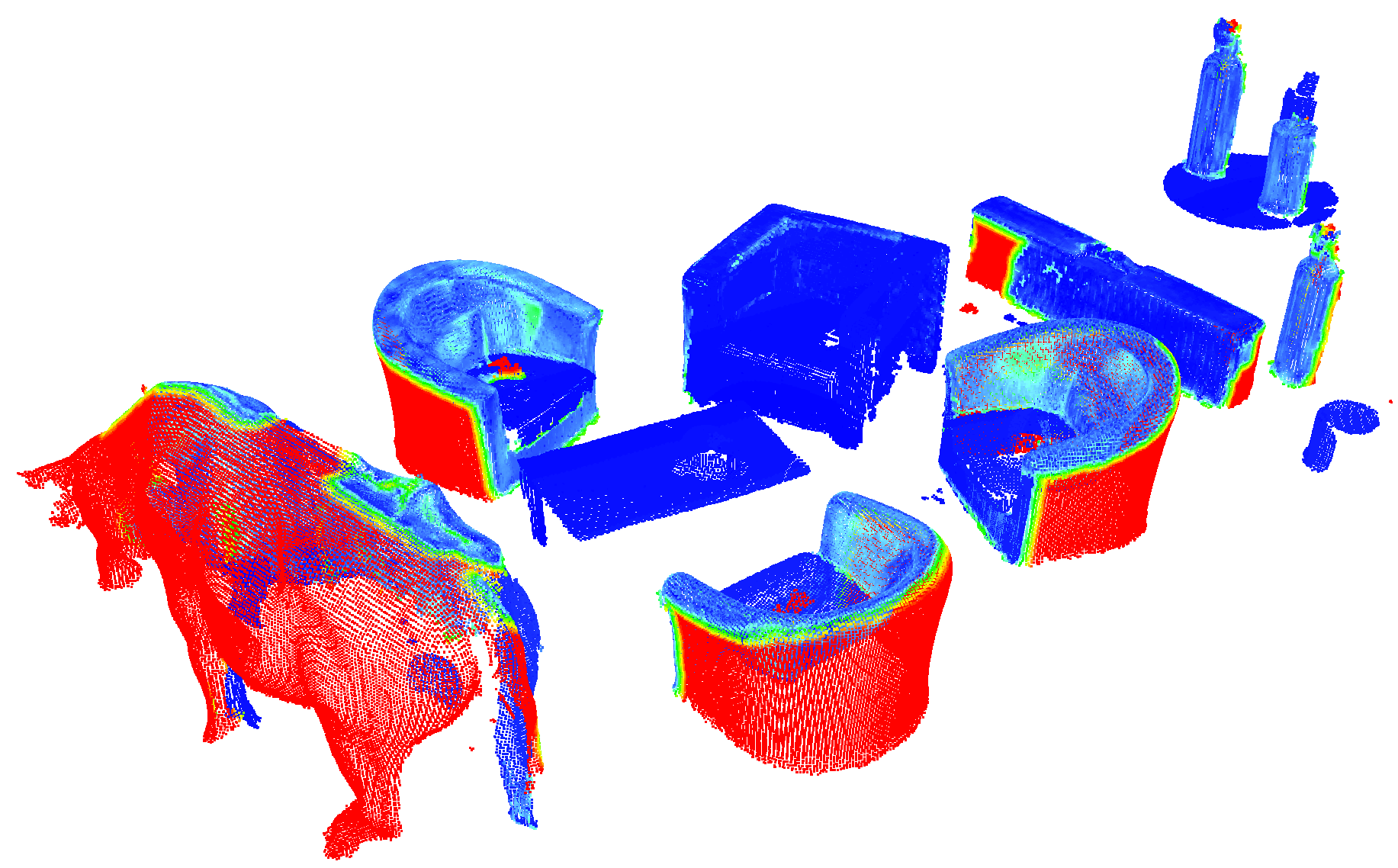}
\caption{Inserting the completed object model instances into the reconstructions results in a more complete and accurate model of the scenes. The images show distances of the points from the completed reconstruction to the points on the originally observed scene (\textit{red:} high to \textit{green:} low). The original reconstructions are shown at the top of Fig.~\ref{fig:scenenn_066_gsm} and the bottom scene of Fig.~\ref{fig:tango_gsm}.}
\label{fig:tango_merged_and_unmerged_raw_distance_to_original}
\end{figure}

Finally, we demonstrate the performance of our system in completing scenes by using detected merged instances of objects in single- and multi-session applications.
The difference between these applications is that in the first case, our system starts with an empty database, while the second setup operates with an existing database (from the previous sessions, which gets extended).
In Fig.~\ref{fig:tango_merged_and_unmerged_raw_distance_to_original}, we depict two scenes in which multiple object models were detected and inserted, one from the SceneNN66 and one from the Tango dataset.
During inclusion, only models that align well with the \acronym{tsdf} grid of the scene are included.
Note, how the object model inclusion succeeds to accurately align with the partial views in the scene (\emph{blue}) and manages to fill gaps of unobserved parts (\emph{red}).

\subsection{Limitations}
We have shown that our system can perform well on different datasets, however some challenges still remain:
\begin{itemize}
    \item \emph{Planar structures and geometrically uninteresting objects}, such as cylinders and cubes are not descriptive enough when partially observed and are, therefore, very hard to merge. Thus, we exclude planar objects from matching. One potential solution to this problem could be to include the visual appearance into the description.
    \item Objects with \emph{thin surfaces} are very hard to represent by a \acronym{tsdf} because of the truncation distance. This is an inherent problem of the representation and can be avoided by using a different one, such as point clouds, meshes, etc.
    \item In the segmentation step, we make that assumption that object can be represented with convex regions, although this is not always the case. In order to deal with \emph{non-convex shapes}, different segmentation techniques could be applied.
    \item For some object models, \emph{descriptors} and \emph{keypoints} were not descriptive enough, failing to properly match even if the same instance of the object in the scene was observed. An alternative would be to use more descriptive features, such as global descriptors, \acronym{tsdf}-based descriptors or learning based descriptors.
\end{itemize}

\section{Conclusion and Future Work}
In this work, we have presented a novel database system for incremental 3D object model generation.
The presented method is capable of automatically discovering new object models and updating, improving, and completing existing models with new observations.
We showed that the knowledge of merged object models can be used to complete scenes and, hence, improve scene reconstructions. 
No prior information about the objects is required for the entire database generation and updating, facilitating a completely unsupervised object detection scheme.
Finally, we evaluated our methodology on one publicly available and on four newly created RGB-D object datasets.
The latter are released with this publication.

The presented object database expedite several emerging applications.
In robotic navigation, for instance, object based SLAM approaches can make use of this system without prior knowledge or strong assumptions about the objects present in the mapped environment.
This gives rise to broader applicability of such mapping systems.
Furthermore, the segment matching and registration procedures within such a mapping system may be simultaneously used for detection of loop closures.
In future work, we wish to incorporate semantic information to the object models, for a faster matching procedure.
We furthermore believe that systems based on other depth sensing modalities, \eg, laser range finders, can benefit from the proposed object database framework in an equal fashion  as RGB-D.

\section*{Acknowledgment}

This work was partially supported by the Swiss National Science Foundation (SNF), within the National Centre of Competence in Research on Digital Fabrication, and the Swiss Commission for Technology and Innovation (CTI).

\bibliographystyle{IEEEtran}
\bibliography{bibliography}

\end{document}